# Bimanual Motor Strategies and Handedness Role During Human-Exoskeleton Haptic Interaction

Elisa Galofaro*†, *Student Member, IEEE*, Erika D'Antonio†, *Student Member, IEEE*, Nicola Lotti, Fabrizio Patané, Maura Casadio and Lorenzo Masia, *Senior Member, IEEE*

*Abstract* — Bimanual object manipulation involves multiple visuo-haptic sensory feedbacks arising from the interaction with the environment that are managed from the central nervous system and consequently translated in motor commands. Kinematic strategies that occur during bimanual coupled tasks are still a scientific debate despite modern advances in haptics and robotics. Current technologies may have the potential to provide realistic scenarios involving the entire upper limb extremities during multi-joint movements but are not yet exploited to their full potential. The present study explores how hands dynamically interact when manipulating a shared object through the use of two impedance-controlled exoskeletons programmed to simulate bimanually coupled manipulation of virtual objects. We enrolled twenty-six participants (2 groups: right-handed and left-handed) who were requested to use both hands to grab simulated objects across the robot workspace and place them in specific locations. The virtual objects were rendered with different dynamic proprieties and textures influencing the manipulation strategies to complete the tasks. Results revealed that the roles of hands are related to the movement direction, the haptic features, and the handedness preference. Outcomes suggested that the haptic feedback affects bimanual strategies depending on the movement direction. However, left-handers show better control of the force applied between the two hands, probably due to environmental pressures for right-handed manipulations.

## I. INTRODUCTION

"CYBERSPACE. A consensual hallucination experienced daily by billions of legitimate operators, in every nation" [1]. The idea of interacting with objects and environments in a different dimension raised up between 1970s and 1980s from the mind of few visionaries that imagined creating a virtual space in which they could have become divinities. With the advancement of technology, virtual reality (VR) employed a major role in several applications, spacing from the entertainment to the rehabilitation and training realms [2]–[6]. However, presently, "*immersive*" devices only involve the visual and auditory systems [7], providing the user with a dissonance that can result in what it is called "cybersickness" [8]. A partial compensation of this problem is the inclusion of the *somatosensory system* in the virtual world, beguiling the body to be in the real world: indeed, in our daily living, the objects' manipulation involves the so-called *haptic sense* [9].

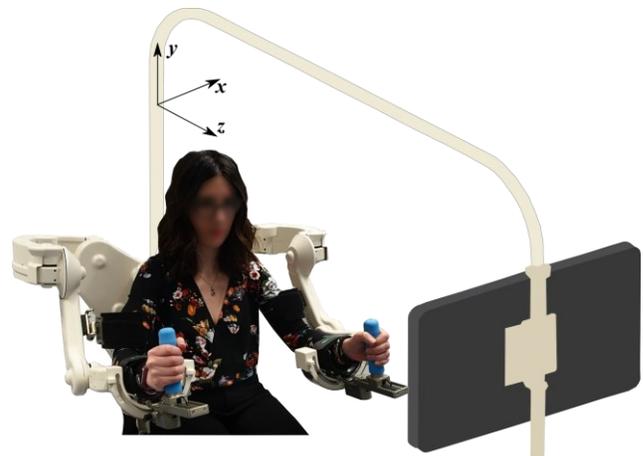

Fig. 1. ALEx-RS robotic device (http://www.wearable-robotics.com/kinetek/). Subject is performing the experiment while wearing the exoskeleton.

Over the last decades, the most common approaches adopted to deliver haptic illusion rely on vibrotactile [10], or cutaneous electro-tactile stimulations [11] or with robotic force fields [12]. Several researchers aimed at developing haptic control and devices able to provide users with tactile and interaction force information [13][14]. However, most of the existing haptic devices consist of desktop systems or end-effector devices with limited workspace and low force rendering capabilities [15][16], without focusing on bimanual actions. Yet, the most common Activities of Daily Life (ADLs) require coordinated use of both arms: a simple task like unscrewing a cork from a bottle and drinking its contents requires the functional coordinated interaction between the two hands in such a way that limbs exploit the bimanual coordination. In such a simple task each hand has its own well-defined role: the non-dominant one stabilizes to steadily hold the object, while the dominant hand finely manipulates it and complete the action. Most of everyday situations requires the recruitment of both arms, yet mechanisms underlying bimanual coordination are still a scientific debate, and the formalization of dynamic laws describing mutual interaction between limbs cannot be generalized across the vast manifold of human dexterity.

† means equal contribution.
Elisa Galofaro (*corresponding author: elisa.galofaro@ziti.uni-heidelberg.de), Erika D'Antonio, Nicola Lotti and Lorenzo Masia, are with the Assistive Robotics and Interactive Exosuits Lab, Institut für Technische Informatik (ZITI), University of Heidelberg, Heidelberg, Germany.
Maura Casadio is with the Department of Informatics, Bioengineering, Robotics, and System Engineering of the University of Genoa, Italy.
Fabrizio Patané (fabrizio.patane@unicusano.it) is with M3Lab Engineering Department of the University Niccolò Cusano, Rome, Italy.

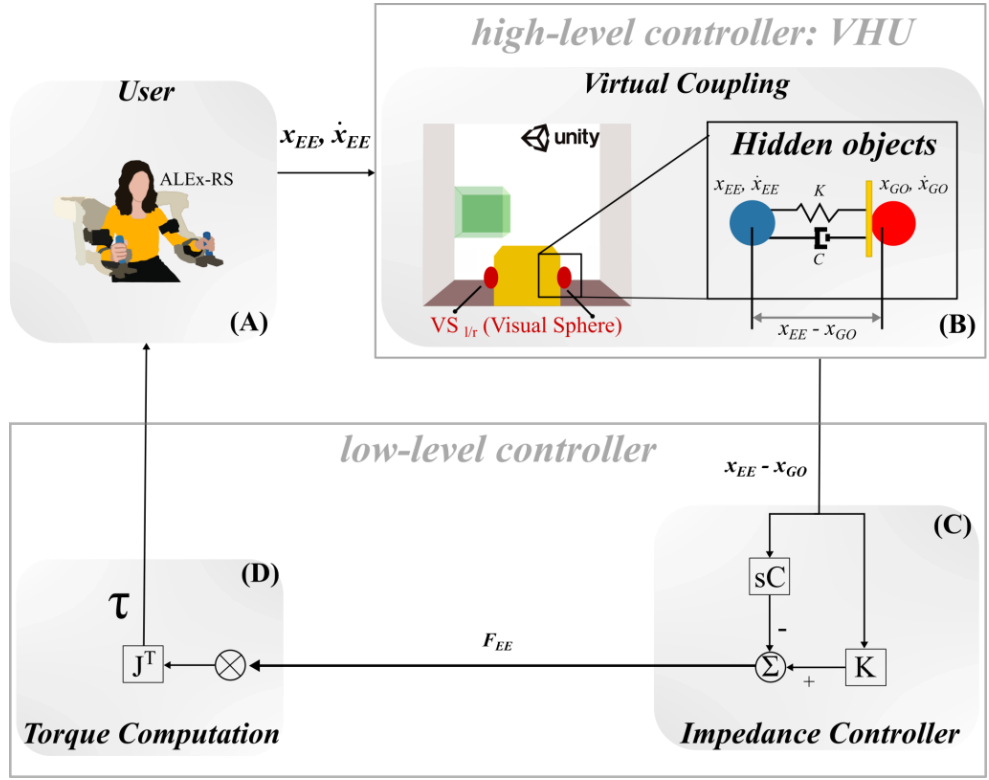

Fig.2. Control Architecture Scheme. **(A)** The user wearing the exoskeleton ALEx-RS. **(B)** High-level controller (Virtual Haptic Unit - *VHU*): a visual representation of the interactive scenario. Here is computed the "physical-based" response to the user's movements in the virtual environment, generated by using Unity 3D. **(C)** Impedance Controller module: this block computes the haptic feedback. The force generation is calculated by means of elastic ($K$) and viscous ($C$) components, which are proportional to the distance between the End Effector spheres ($x_{EE}$: $x_{EEl}$ or $x_{EEr}$) and the God-Objects ($x_{GO}$: $x_{GOl}$ or $x_{GOr}$) and the corresponding velocities of the End Effector spheres ($\dot{x}_{EE}$: $\dot{x}_{EEl}$ or $\dot{x}_{EEr}$) and the God-Objects ($\dot{x}_{GO}$: $\dot{x}_{GOl}$ or $\dot{x}_{GOr}$). **(D)** Torque Computation module: here the force values at each EE ($F_{EEr}, F_{EEl}$) are then sent back to the device, and then converted in torques at the joints ($\tau$) of each exoskeleton, by means of the transpose of the Jacobian ($J^T$).

Literature offers a noteworthy number of studies involving bimanual haptic settings targeting multiple purposes: from rehabilitation to general motor control and haptics. These contributions and the employed setups have been persuasively summarized in a review paper by Talvas and colleagues [17]. *Bimanual task* could be categorized in two distinct types of interaction: (i) *uncoupled bimanual task*, where hands separately act without a common objective and on separate workspaces [18]–[20], (ii) *coupled bimanual task* in which they mutually interact with a common purpose, i.e., by manipulating the same object concurrently [13], [21], [22]. Mechanisms underlying the control of bimanual actions have been extensively investigated for *uncoupled* tasks [18]–[20], while few contributions specifically focused on bimanual tasks in *coupled* settings [13], [21], [22]. Research mainly focuses on studying *uncoupled schemes* with object manipulation exposed to gravity and various artificially generated force fields [23]–[25]. The few investigations on *coupled bimanual* tasks followed a different approach by focusing on the person's handedness. Two theories exist which describe the mechanisms underlying the interconnection of dominant and non-dominant hands during bimanual manipulation: (i) the coordinated use of limbs is highly weighted by the dominance hand, which primarily acts (imposing the task dynamics) while the other stabilizes the manipulation [26]–[29]; (ii) the hands switch their functional role across various environmental constraints and task difficulties [21], [30]. For example, in a study involving *coupled* bimanual object manipulation through a dual-wrist robotic interface, Takagi and colleagues [31] tried to characterize the role of each limb: conversely to the dynamic-dominance theory [32], the authors found that subjects preferred to stabilize the manipulated object with the dominant hand. They extracted such information analyzing the wrist muscular activity, which revealed a preference for co-contracting the dominant hand during both object holding and transport. Another study had also led to similar considerations: Woytowicz et al [33], by employing a physically *coupled* bimanual task where the hands of right-handed subjects were coupled together by a spring using a planar manipulandum.

However, the few studies discovered in the literature that investigate *coupled bimanual* tasks [31], did not rely on recent advancements in haptics and robotics that have the unexplored potential to provide realistic scenarios involving the whole upper limbs during multi-joint movements. The inclusion of such variables, as well as the inclusion of left-handed subjects, could highlight unpredictable kinematic strategies, highly dependent on the environmental context and society constraints (i.e., objects that are strictly designed for right-handed people).

Hence in the present study, we implemented a virtual reality scenario and a specific task where subjects performed a coupled manipulation while wearing a bimanual exoskeleton (ALEx-RS, Wearable Technology, Italy): right and left-handed participants were requested to grab and lift virtual objects using two arms and move it across a 3D-workspace towards multiple target positions. Different visuo-haptic feedback conditions were implemented to replicate objects of various dynamics and, therefore, compliances with associated weight and inertia.

The central hypothesis was that an opportunely designed combination of haptic and visual feedbacks would affect manipulation strategy in a three-dimensional task. Based on previous evidence [34], [35], we also hypothesize that the hand dominance might influence the bimanual motor strategy during manipulation and target reaching.

Considering the importance of *haptic feedback* and *coupled bimanual interaction* for practical applications and the available robotic technologies, the purpose of our contribution is to provide evidence that our *setup* could be a further step toward a more realistic virtual object manipulation scenario to study human motor strategies. Thus, the purpose of our study is threefold: (i) to implement an exoskeleton-based haptic interface with virtual objects with different mechanical impedance; (ii) to identify the strategies of motor control in bimanual manipulation, and how the handedness influences them; (iii) to characterize how these motor strategies change according to the perception of different simulated objects with variable compliances.

The paper is organized as follows: Section II describes our experimental setup. Section III presents the real-time control architecture that includes three main components: a ***Virtual-Haptic Unit*** (***VHU***), an ***Impedance Controller***, and a ***Torque Computation module***. Then, Section IV explains the task implementation and the experimental protocol. Section V provides details on the outcome measures and statistical analysis. Section VI presents the results which demonstrate our hypothesis. Section VII discusses the meaning of our results in terms of bimanual manipulation strategies and motor performance. Here, we also discuss the benefits and limitations of our approach. Finally, Section VIII concludes and details several future research directions.

## II. EXPERIMENTAL SETUP

The experiments have been carried out using the Arm Light Exoskeleton ALEx-RS [36]–[38] (Figure 1): the device consists of a bimanual exoskeleton for upper limbs assistance, mobilizing six degrees of freedom (DoFs) per arm. Four DoFs are sensorized and actuated: shoulder abduction-adduction (sh-AA), shoulder pronation-supination (sh-PS), shoulder flexion-extension (sh-FE) and elbow flexion-extension (eb-FE). The remaining DoFs, i.e., wrist pronation-supination (wr-PS) and wrist flexion-extension (wr-FE), are only sensorized, and they did not provide assistance to the user. The actuated joints are powered by 4 brushless motors providing maximum torque values of 35 Nm (sh-AA), 35 Nm (sh-PS), 25 Nm (sh-FE), 20 Nm (eb-FE). Optical incremental encoders are integrated into the motor groups, and absolute angular sensors are directly integrated into the joints to measure angular rotations. The robot can be controlled by specifying the desired forces at the end-effector (EE), which are then translated into torques at each actuated joint.

The exoskeleton can cover 92% of human arms' range of motion (ROM) and, due to its tendon-driven transmission for torque delivering, the structure is extremely light [39].

The high-level control includes the possibility to use the workstation in three active modalities, which can be defined with respect to the robot action: i) "*gravity compensation*", in which the user moves both the arms in a back-driveable mode, ii) "*assistive*", in which the robot drives the limbs during the task execution, and iii) "*assisted-when-needed*", the exoskeleton guides the user's arm when the movement has not started after a predefined time. The device also provides gravity and friction compensation. The inertia is partly [40] cancelled by an inverse dynamic model that online runs in the background and derives the user's motion from the current absorbed by the motors.

The haptic control framework has been wholly re-programmed in a new bimanual modality able to provide torques at the joints when the user interacted with simulated objects in a virtual environment. All the details about haptic computation are discussed in the following.

## III. REAL-TIME CONTROL ARCHITECTURE

The controller's architecture has been designed to provide stable force feedback during manipulations of virtual objects'

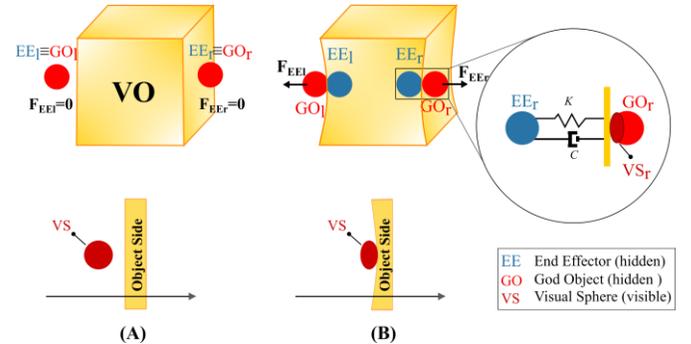

Fig. 3. Principle of the *Virtual Coupling God-Object algorithm*: although the haptic EEs (blue spheres) penetrate the virtual surfaces, the god-objects (GOs, red spheres) are constrained to remain on the surface of the obstacles. **(A)** Representation when the GOs and the EEs are in the surrounding space without touching obstacles, they are exactly overlapped, and no force is provided to the user. **(B)** Representation when the EEs compenetrate the VO and the GOs remain on the surface providing forces proportional to the distance between the GOs and the EEs. The right side shows the deviation that is created between EE and GO when the EE enters the VO making possible the haptic feedback resulting from this distance and the elastic and viscous components. The *visual* feedback provided to the user during the experiment is represented only by the right and left visual spheres ($VS_{l/r}$). The two elements used to compute the haptic feedback (EE and GO) remain hidden during the entire task. The bottom line simplifies the description by setting out in 2D what is represented in the line above.

and included three main components (Figure 2): a ***Virtual-Haptic Unit*** (***VHU***), an ***Impedance Controller***, and a ***Torque Computation module***.

*A. Virtual Haptic Unit - VHU*

The ***VHU*** (Figure 2B), developed in the main desktop workstation via c# code, provided the visual representation of the interactive scenario, and it computed the "physical-based" response to the user's movements in the virtual environment, generated by using Unity 3D ([41], version 2019.2.13f1). This module allowed providing the collision detection between the virtual counterpart of the handles' device and the VE: the resulting data were sent in real-time as input to the ***Impedance Controller*** (Figure 2C) at a frequency of 200 Hz to provide the haptic feedback.

*A. Impedance controller*

The ***Impedance Controller*** (Figure 2C), running at 1 kHz on a second dedicated PC, computed the haptic feedback providing the interaction force at each EE. The position of each EE in the ***VHU*** was calculated by means of the forward kinematics using the absolute angular positions provided by the sensors integrated into the exoskeletons' joints. The EEs three dimensional coordinates (X,Y,Z) were then mapped into the virtual scenario by using a 2:1 linear scale between the robot motion and the virtual reality displacement. Haptic rendering was implemented using the aforementioned "*God-Object method*" that included friction simulation [42]: it employs two representations of each EE to implement an impedance control rendering, through the stiffness and damping generated during interaction with the simulated object (***Impedance Controller***). Such an approach is called "Virtual Coupling" and mediates the interaction between the user and the simulated environment.

In details, each hand position is represented in the ***VHU*** by three spheres, respectively named End Effector (EE), God Object (GO), and Visual Sphere (VS), and they are used for different computation or visualization purposes (Figure 3):

- A blue sphere (End Effector, EE), not visible to the user, represents the end point used to compute the haptics and corresponds to the real EE position in the virtual scenario.
- A red sphere, named God Object (GO), not visible to the user, feeds back the EE position in the VR and deformation due to contact with the VO.
- An additional Visual Sphere (VS), visible to the user, feeds back the EE position in the VR and deformation due to contact with the VO. This sphere is not used to compute the forces, but only the *visual feedback* to complement the *haptic feedback* [13]. Also, its form factor is modified to provide contact information (visual deformation).

For ease of representation, only the two spheres responsible for the haptic feedback are depicted in the top side of Figure 3 (EE and GO). Considering Figure 3A, in the absence of contact between the EEs and the VO (or the room walls), the two spheres (blue and red) move synchronously, and no force is fed back to the user. Since the EE does not have the "physics collider" property [43], the generation of haptic feedback is computed when it enters the VO volume. The blue spheres in Figure 3 ($EE_l$ or $EE_r$ for left or right end effectors, respectively), followed the actual position of the EEs of the exoskeleton, which are penetrating the object, while the GO positions ($GO_l$ or $GO_r$, red spheres) remains tangent to the object (thanks to its "physics collider" properties) Figure 3B. Generation of forces is progressively calculated, based on the penetration of the EE inside the VO, in the ***Impedance Controller*** module by means of elastic repulsive components, which are proportional to the distance between the End Effector spheres ($EE_l$ or $EE_r$) and the God-Object ($GO_l$ or $GO_r$), according to the scheme depicted in Figure 2B-C. Simultaneously, a viscous term is provided by the mutual velocity between the two spheres during motion [44]. To provide visual information of the contact force, the VSs are deformed by changing the shape factor using the EE's penetration depth into the VO, Figure 2C. The spheres are squeezed on the object surface proportionally to the applied forces and, therefore, to the distance between EE and GO.

The haptic feedback to the user has been provided for each EE by computing the interaction force $\vec{F}_{EE}$ along the three dimensions of motion:

$$\vec{F}_{EE} = \begin{bmatrix} -\Delta X \cdot K_0 \cdot (1-\varepsilon) - \Delta \dot{X} \cdot C \\ M/2 \cdot g \\ 0 \end{bmatrix}, \quad (1)$$

where $M = 0.2 \, kg$ is the mass of the virtual cube, $g$ is the gravity acceleration, $K_0 \cdot (1-\varepsilon)$ represents the object's stiffness where $K_0$=150 N/m, and $0 \leq \varepsilon \leq 1$ is the parameter manipulated for deforming the VO (see below), $C = 5 \, Ns/m$ is the damping ratio of the viscous force, $\Delta X$ is the distance between the *EE* position ($x_{EE}$) and the *GO* position ($x_{GO}$), and $\Delta \dot{X}$ is the difference between the *EE* ($\dot{x}_{EE}$) and the *GO* ($\dot{x}_{GO}$) linear velocities obtained by discrete-time differentiation.

VO parameters were chosen considering the haptic rendering capability of the device for not incurring in saturation during the task.

Since the task required to lift the VO using the two EEs, we also simulated the vertical static friction ($F_f$), which assures contact between the EEs and the object preventing slippage by generating a minimum *contact force*. The dynamics of the VO, $\vec{F}_{cube}$, depends on the forces applied by the EEs during bimanual manipulation ($\vec{F}_{EEr}$ and $\vec{F}_{EEl}$), on the gravity force $M * \vec{g}$, on the static friction ($F_f$), and on the acceleration ($\ddot{x}$, $\ddot{y}$, $\ddot{z}$,) and was described by the following equation:

$$\vec{F}_{cube} = \begin{bmatrix} M\ddot{x} + F_f \\ Mg + M\ddot{y} \\ M\ddot{z} \end{bmatrix} + \vec{F}_{EEr} + \vec{F}_{EEl} \quad (2)$$

Since the virtual coupling algorithm, developed on Alex-RS, requires a punctual interaction between the robot end-effector and the virtual object to generate a stable interaction and avoid undesired and uncontrolled movements, the VO has been manipulated only in a three and not six-dimensional manifold [13], [17], [31]. Linear movements along the orthonormal axes (XYZ) were allowed, but rotations ($\theta_X$, $\theta_y$, $\theta_z$) were disabled. Forces and relative haptic feedback were computed using single interaction points between the object's surface and the EEs, which makes it impossible to balance rotations around the coordinated axes unless forces are directionally aligned and with opposite magnitude. Hence, the VO can be moved along the coordinated axes but not rotated, providing across the workspace reaction forces generated by the interaction with the EEs.

### B. Torque computation module

The force values at each EE ($F_{EE_r}, F_{EE_l}$) are then sent back to the device through a shared memory communication protocol, and then converted through the **Torque Computation** module (Figure 2D) in torques at the joints ($\tau$) of each exoskeleton, by means of the transpose of the Jacobian $J$:

$$\tau = J^T * F_{EE_{r/l}} \quad (3)$$

The different objects were implemented by modelling the two following *virtual physical properties*, as depicted in figure 4:

1. *Stiffness* ($K_m$): by implementing the following relationship:

   $$K_m = K_0(1 - \varepsilon_m), \quad (4)$$

   Where the index $m=1, \ldots, 4$ indicates one of the four implemented material, characterized by $\varepsilon_m$.

2. *Breakage limit* ($F_{break}$): by varying the breaking point in terms of maximum force that the material could withstand without collapse and resulting in a failed trial. This breakage limit was introduced for safety reasons when using the exoskeleton and to avoid damage to the same.

All the low-level procedures were performed in Matlab/Simulink® environment (MathWorks, Natick, Massachusetts MA, USA).

## IV. HAPTIC FEEDBACK VIA ROBOTIC EXOSKELETON

### A. Task implementation

The scenario replicated a virtual room enclosed by four walls (3 walls and the floor) in which are visualized: (i) the virtual object (VO), represented as a cube, (ii) the position of the device end-effectors (EEs), resembled as spheres, and (iii) the final target location, designed as square-shaped portion highlighted in the workspace.

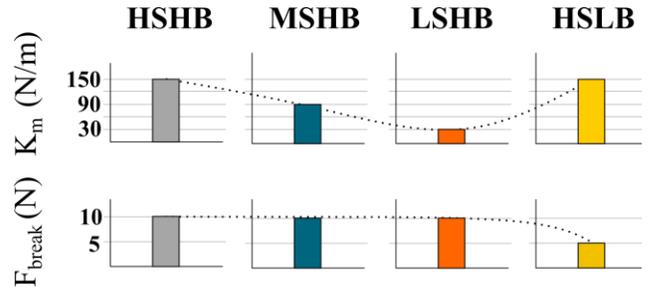

Fig.4. Physical features of the four virtual objects (VOs): **HSHB** (High Stiffness, High Breaking point), **MSHB** (Medium Stiffness, High Breaking point), **LSHB** (Low Stiffness, High Breaking point) and **HSLB** (High Stiffness, Low Breaking point).

The task based our previous preliminary study [45], consisted of bimanually grabbing and lifting the VO, securing it in a stable interaction without exceeding force thresholds (Figure 5A), and successively delivering it to an indicated target position (Figure 5B).

During the pick and place task, an appropriate level of contact force between the *EEs* and the *VO* should be kept to prevent failure, which can occur in two different ways: (i) when a sufficient level of *Contact Force* along the x-axis ($F_{contact}$) was not maintained during grabbing the *VO* with the two *EEs* hence causing the object slippage and falling; (ii) when exceeding grabbing force ($> F_{break}$) causing the object breakage, an aspect that is extensively addressed in previous studies [46],[47].

We simulated different haptic renderings by combining four gain sets corresponding to objects with different impedance behaviours in terms of stiffness and damping, while the mass was left constant. Objects' textures were chosen to provide users with visual information on the "material" to manipulate, Figure 4:

- "HSHB – *High Stiffness, High Breaking point*" in which subjects were provided with a rigid-looking object depicted with metallic surfaces, characterized haptically by high stiffness: the object was difficult to deform, presenting high rigidity and the $F_{break}$ was set high.

- "MSHB - *Medium Stiffness, High Breaking point*": subjects experience a compliant object which can be deformed yet presenting a medium/high stiffness. The visual texture resembled elastic components, and the breakage point was set high.

- "LSHB – *Low Stiffness High Breaking point*": the object was characterized by a very low stiffness, and therefore was highly deformable while the breaking threshold was set high. The visual texture resembled soft components.

- "HSLB- *High Stiffness, Low Breaking point*": this last

haptic simulation aims at replicating a fragile object, as suggested by the visual texture. The stiffness was set to high, and yet the breakage threshold was low. Subjects did not experience compliance.

The specifications of the parameter for four different VOs are presented in the Table I.

*Table I  Virtual Objects Properties*

|  | **Virtual Objects** | | | |
| --- | --- | --- | --- | --- |
|  | **HSHB** | **MSHB** | **LSHB** | **HSLB** |
| $F_{break}$ (N) | 10 | 10 | 10 | 5 |
| $\varepsilon_m$ | 0 | 0.4 | 0.8 | 0 |
| $F_{contact}$ (N) | 2 | 3 | 3 | 2 |
| $k_m$ (N/m) | 150 | 90 | 30 | 150 |

*B.  Subjects*

A group of twenty-six healthy young subjects (11 males and 15 females, 24.7±3.9 (mean ± std) years old, range: 20-33 years), took part in the study. Within the group, there was no significant difference in the age distribution between males and females; fifteen subjects were right-handed according to the Edinburgh Handedness Questionnaire [48] (Laterality score (LS) = 83.78 ± 15.91 (mean ± std)). Eleven participants were left-handed (Laterality score (LS) = -63.64 ± 22.92 (mean ± std)). All participants provided their informed consent before the experiment, and the experimental protocol was approved by Heidelberg University Institutional Review Board (S-287/2020): the study was conducted following the ethical standards of the 2013 Declaration of Helsinki. Experiments were carried out at the Aries Lab (Assistive Robotics and Interactive Exosuits) of the Heidelberg University. Subjects did not have any evidence or known history of neurological diseases and exhibited a normal joint range of motion and muscle strength.

*C.  Experimental task and procedure*

Participants sat on a comfortable chair equipped with adjustable footrests in front of a 43" monitor on which the virtual scenarios were shown. The user's arms and forearms were secured by means of Velcro straps and safety belts on the exoskeletons, Figure 1.

For every single trial, participants were requested to grab the object from the lateral sides, lift, transport and place it on the target (reach phase) from a predefined starting point.

Three target positions (left, L=-45°; centre, C=0°; right, R=45°) were presented in order to study and analyse the motor control strategies across different positions in the workspace, Figure 5A-B. Targets were placed along a semicircle, equally spaced from the starting position (radius = 40 cm$_{robot}$, depth =15 cm$_{robot}$): each target was presented six times, obtaining a total of 18 repetitions for task (material condition). The targets were proposed as yellow squared shape pictures with the same dimensions of the VO along the x- and y- axes. A semi-transparent green squared shape was positioned in the same x and y coordinates of the target, but at a distance equal to the size of the cube along z-axis, in order to help subjects in understanding the environment perspective, since we represented a virtual 3D scenario on a 2D screen. No time constraints were imposed to avoid participants' anxiety and their inability to complete the task. An *auditory cue* informed subjects to stop moving the cube along z-axis when they arrived in the right position on the target. After that, subjects were instructed to start with the new trial.

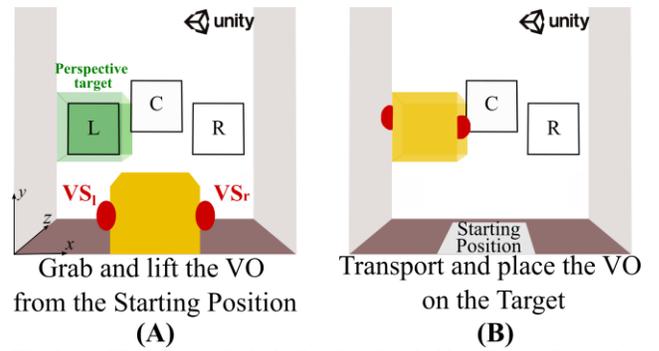

Fig. 5. **(A)** Virtual scenario including the virtual object (VO, yellow cube) and the god-objects (GO$_l$ and GO$_r$, red shapes) during the initial phase of the task. **(B)** Virtual scenario during the reaching of the left target (L).

Before starting the experiment, a familiarization phase was performed. Participants were instructed to maintain an appropriate level of contact force with the lateral sides (along the plane x-y) of the VO throughout all trials and not incur failure, Figure 5a. In case of failure, an *auditory cue* was provided based on a specific case: one for the breaking and another one for the falling condition. After the cue, subjects had to interrupt the task and return to the starting point for a new trial.

The experiment was organized in a one-day single session, and four different conditions have been randomly assigned to each participant, with a 5-mins rest break between different conditions. The whole session had a total duration of about 1 hour. A priori experiment to evaluate the exoskeleton transparency was not performed since we rely on previous experimental evidence [37], which shows how Alex-RS, programmed in different modalities, generates kinematics and muscle coordination comparable to natural movements.

## V.  Data Analysis

Forces and trajectories were recorded at 200 Hz and offline filtered using a 6$^{th}$ order low-pass Butterworth filter with a 10 Hz cutoff frequency. We considered geometric and kinematic factors to extract the dynamics experienced by the participants and extrapolate the indicators for characterizing their performance.

*A. Outcome Measures*

The **Deflection** parameter indicates the quantitative differences between the four implemented virtual objects, it is related to the haptics simulation: it represents the lateral deformation of the objects caused by the interaction of the user's hands with the object itself. Its values were expected to decrease when the stiffness (or breaking point - HSLB vs. HSHB) increased. The **Deflection (m)** of the VO (Figure 6A) during the interaction with the end-effectors of the dual exoskeletons was the primary data: defined as the difference between the VO initial width ($VO_{\text{initial width}}$ before contact with the EEs) and the Euclidean distance (along the x-axis) computed between the EEs ($x_{EE_l}, x_{EE_r}$):

$$Deflection = VO_{\text{initial width}} - \sqrt{(x_{EE_l} - x_{EE_r})^2} \quad (5)$$

Regarding subject' performance measures: for each target, data were collected and successively analyzed, considering both *successful trials* and *failure trials*. The *successful trials* were defined as those where the object was successfully placed on the target, without any breakage or slippage. The *failure trials* were those where participants applied a too large grabbing force of undershooting the minimum contact force, provoking object slippage and consequent breakage.

We computed the following indicators using only *successful trials*, by post-processing the end-effector trajectories in both spatial and temporal domains.

The **Dynamical Symmetry Index (DSI, %)**, as defined in [13], [49], [50] evaluates the temporal coordination during coupled bimanual manipulation tasks, and it is computed as the percentage difference between the trajectories of the two EEs across the trials:

$$DSI(t) = \frac{d_r(t) - d_l(t)}{d_r(t) + d_l(t)} * 100 \quad (6)$$

where $d_r(t)$ and $d_l(t)$ are the distances between the initial grabbing point (computed as the point in which $F_x >$ minimum contact force) and the instantaneous position of the dominant and non-dominant hand, respectively. This is an indicator of bimanual coordination in object manipulation [13]. It ranges from -100% to 100%: positive values indicate that the right EE trajectory is longer than that of the left EE, while negative signs of the **DSI** describe an opposite situation in which the non-dominant EE traces a longer path than the dominant one. A value of 0% reflects perfect symmetry. **DSI** values between ±5% indicate symmetry since right EE trajectory is almost equal to the one of the left EE. Larger values are considered indicating the presence of asymmetry between the two limbs.

The **Normalised Jerk (NJ)** [51], [52] has been considered as an indicator of movement smoothness, and it is calculated by the following expression:

$$NJ = \sqrt{\frac{T^5}{2L^2} \int_0^T jerk(t)^2 \, dt} \quad (7)$$

where $T$ is the execution time for a single trial, $L$ is the Path length and $jerk(t)$ is the jerk index equal to the time derivative of acceleration, meaning the third derivative of the EE trajectory ($jerk(t) = \dddot{x}(t)$, $x$ = EE trajectory), [53]. The lower the values of this metric, the better the smoothness of movement and thus the subjects performance. **NJ** is normalized with respect to execution time and path length such that trajectories of different duration and size can be compared.

To observe the change in manipulation strategies across subjects, we evaluate the **Force Profile** shapes resulting from the haptic interaction between the end effectors $EE_{r/l}$ and the **VO**, given by the equation (1). In particular, we focused such analysis on the interlimb differences of forces applied by the $EE_{r/l}$, which indicates the variation of coordination between the two hands [54]. We normalized the **Force Profile** with respect to the total length of the trajectory performed from the initial grab of the **VO** to the successfully reaching of the target.

The following metrics are extracted from data referred to both the *successful* and *failure trials*.

**Force Percentage Break (FPB, %)**, computed as the ratio between the total force applied on the $EE_l$ ($F_{EE_l}$) and the sum of the force on both the **EEs** ($F_{EE_r}$ and $F_{EE_l}$) in the instant in which subjects exceeded $F_{break}$:

$$FPB = \frac{F_{EE_l}}{F_{EE_l} + F_{EE_r}} * 100 \quad (8)$$

This is an indicator of the hand responsible for breaking the **VO**. It ranges from 0% to 100%: values between 0% and 50% indicate that the dominant hand exceeded the imposed force limits, while values between 50% and 100% describe an opposite situation in which the non-dominant hand exceeded the force limits.

Finally, we computed the subsequent subject' performance related indicators using even *failure trials*.

The **Execution Time (s)** is defined as the amount of time taken to successfully move the object from the starting position to the target, including the failure trials.

Finally, the **Maximum Failure (MF)** is computed as the maximum times of the **VO** breaking event during all the experiments for every condition.

### A. Statistical analysis

The metrics **Deflection**, **DSI**, **NJ**, **Force Profile**, and **FPB** were averaged over time of every single trial. We used a repeated-measures analysis of variance (rANOVA) on the dependent variables. We considered as within-subjects factors: '*Target*' (Right (R), Central (C), Left (L)), '*Material*' (HSHB, MSHB, LSHB, and HSLB) and '*Hand*' (left hand (LH), right hand (RH)) for the metrics **Force Profile** and **FPB**; regarding instead the metrics **DSI**, **NJ**, **Execution Time**, and **MF** we considered as within-subjects factors only '*Target*' and '*Material*'; finally we included only '*Material*' factor for the **Deflection** metrics.

Data normality was evaluated using the Shapiro-Wilk Test and the sphericity condition was assessed using the Mauchly test. Statistical significance was considered for p-values lower than 0,05. Post-hoc analysis on significant main effects and interaction was performed using Bonferroni corrected paired t-tests. Statistical analysis was conducted by using IBM SPSS Statistics 23 (IBM, Armonk, New York, USA).

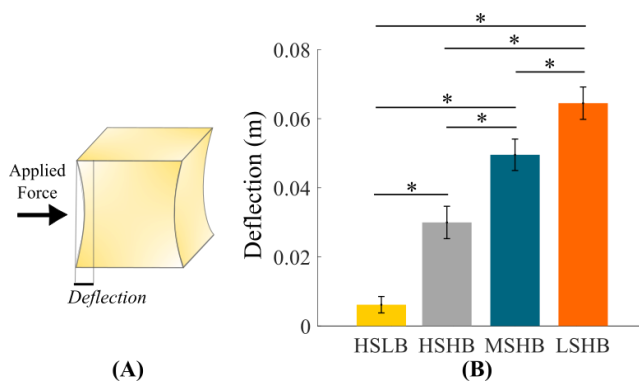

Fig.6. **(A)** Schematic description of the *Deflection* metrics: the red line evidences the effective measurement. **(B)** Deflection metrics for the four implemented virtual objects on ALEx RS. Each bar represents the mean value and the respective SE.

## VI. RESULTS

### A. Discrimination between the implemented virtual objects

We evaluated the **Deflection** parameter for each simulated object to have a preliminary view of the manipulation strategies across the different conditions and across multiple trials. The results are illustrated in Figure 6B, which shows, as expected, that higher deflection during manipulation happens for softer materials. From the statistical analysis with rANOVA we highlighted an effect of the material ('*Material*' effect: F=380.42, p<0.001). The post-hoc analysis following this significant main effect highlighted a significant difference between pairs of objects with different mechanical proprieties (post-hoc: p<0.001 for all the comparisons).

### B. Dynamic manipulation strategies depend on the leading hand across the workspace

Figure 7A depicts the **force profiles** averaged across all the subjects for each condition, showing each hand contribution (left- and right- hand forces) for the left (L), centre (C) and right (R) portions of the workspace where the targets were placed. They all have a bell-shaped profile, with an initial raising force, a single peak of maximum force and a decrease of force, which are applied to grab, lift, and transport the object, respectively. Yet, evident differences across simulated materials can be qualitatively observed at first glance, which also seems to change trend depending on the portion of the workspace and the direction of motion across trials. Hence, we ran a statistical analysis of the mean force (Figure 7B) finding a significant difference between the materials ('*Material*' effect: F=F=97.108, p<0.001), the direction of movement ('*Target*' effect: F=4.090, p=0.034), the employed hand ('*Hand*' effect: F=37.108, p<0.001), the handedness interaction with the target and the hand ('*Handedness*Target*Hand*' effect: F=9.198, p=0.002) the target interaction with the hand ('*Target*Hand*' effect: F=34.351, p<0.001), and the target interaction with the material ('*Target*Material*' effect: F=2.868, p=0.017) .

In right-handed people, for targets towards the right direction the leading hand is the non-dominant one, vice versa for targets towards the left direction the leading hand is the dominant one. In fact, the post-hoc analysis between the hands showed larger values of the haptic force applied with the left hand only for the right target (post-hoc analysis: p<0.001 for all materials, Table II). Conversely, for the other two target directions (L and C), the post-hoc analysis did not show statistical differences between the two hands (p>0.011 for all materials). When the non-dominant hand plays the leading hand's role, i.e., in movements directed to the right, it cannot modulate the force correctly and then is inclined to exert more force than the strictly necessary. A different result has been found for left-handed people: no statistical differences between the forces applied by the two hands have been obtained in all the target directions (p>0.02), meaning that left-handers present a more coordinate force control when moving within the workspace.

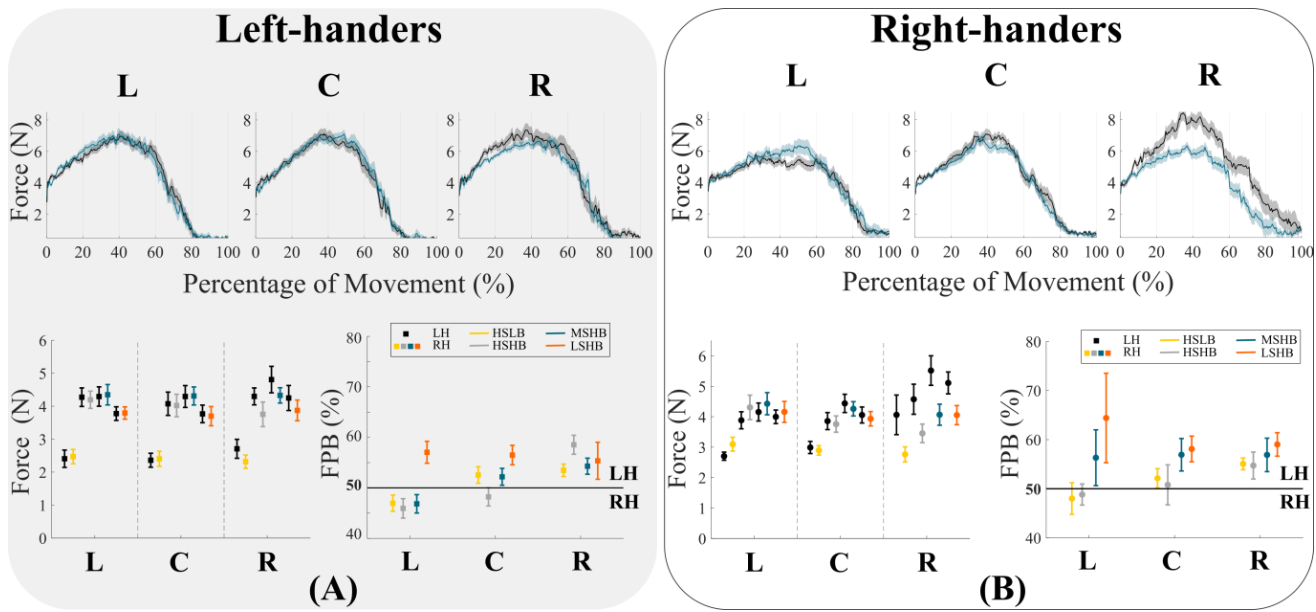

Fig.7. **(A)** Left-handers panel. From top line: normalized force profile trajectories (mean ± SE) relative to the *left hand* (black line) and the *right hand* (blue line) depicted for one sample material (MSHB) and for each target direction (left-L, center-C and right-R). On the x-axis is represented the percentage of movement during the forward movement (from the grasping phase until the object is released on the target). On the left bottom line: averaged force computed across subjects. Mean and SE are highlighted for each material and for each side (*left hand*: black marker, *right hand:* coloured marker). On the right bottom line: *Force Percentage Break* – FPB: mean and SE are highlighted for each virtual object. Values over the black line (FPB > 50 %) highlight the maximal force has been exceeded ($F_{break}$) breaking the object with the *left hand* (LH). Vice versa for the *right hand* (RH) with values under the black line. **(B)** Right-handers panel.

**Table 2.** Mean Force and statistical p-values among the four virtual objects

| VIRTUAL OBJECT | | RH FORCE [N] (Mean ± SD) | | | LH FORCE [N] (Mean ± SD) | | |
|---|---|---|---|---|---|---|---|
| | | L | C | R | L | C | R |
| HSHB | | 4.31 ± 0.98 | 3.75 ± 0.66 | 3.45 ± 0.75 | 3.88 ± 0.68 | 3.86 ± 0.69 | 4.57 ± 1.21 |
| | MSHB | 4.27 ± 0.89 (p = 0.557) | 4.26 ± 0.57 (p = 0.008) | 4.07 ± 0.85 (p = 0.027) | 4.16 ± 0.72 (p = 0.211) | 4.44 ± 0.73 (p = 0.008) | 5.52 ± 1.19 (p = 0.032) |
| | LSHB | 4.15 ± 0.84 (p = 0.540) | 3.93 ± 0.58 (p = 0.425) | 4.05 ± 0.77 (p = 0.015) | 4.00 ± 0.54 (p = 0.511) | 4.06 ± 0.64 (p = 0.305) | 5.11 ± 0.88 (p = 0.104) |
| | HSLB | 3.10 ± 0.56 (p<0.001*) | 2.90 ± 0.38 (p<0.001*) | 2.76 ± 0.61 p=0.001* | 2.70 ± 0.33 (p<0.001*) | 2.99 ± 0.49 (p = 0.001*) | 4.06 ± 1.59 (p = 0.242) |
| MSHB | | 4.27 ± 0.89 | 4.26 ± 0.57 | 4.07 ± 0.85 | 4.16 ± 0.72 | 4.44 ± 0.73 | 5.52 ± 1.19 |
| | LSHB | 4.15 ± 0.84 (p= 0.135) | 3.93 ± 0.58 p=0.039 | 4.05 ± 0.77 p = 0.938 | 4.00 ± 0.54 (p = 0.427) | 4.06 ± 0.64 (p = 0.049) | 5.11 ± 0.88 (p = 0.058) |
| | HSLB | 3.10 ± 0.56 (p<0.001*) | 2.90 ± 0.38 (p<0.001*) | 2.76 ± 0.61 (p<0.001*) | 2.70 ± 0.33 (p<0.001*) | 2.99 ± 0.49 (p<0.001*) | 4.06 ± 1.59 (p = 0.003) |
| LSHB | | 4.15 ± 0.84 | 3.93 ± 0.58 | 4.05 ± 0.77 | 4.00 ± 0.54 | 4.06 ± 0.64 | 5.11 ± 0.88 |
| | HSLB | 3.10 ± 0.56 (p<0.001*) | 2.90 ± 0.38 (p<0.001*) | 2.76 ± 0.61 (p<0.001*) | 2.70 ± 0.33 (p<0.001*) | 2.99 ± 0.49 (p<0.001*) | 4.06 ± 1.59 (p = 0.011) |

**Table 3.** Mean Force and statistical p-values among the four virtual objects

| VIRTUAL OBJECT | | RH FORCE [N] (Mean ± SD) | | | LH FORCE [N] (Mean ± SD) | | |
|---|---|---|---|---|---|---|---|
| | | L | C | R | L | C | R |
| HSHB | | 4.20 ± 0.64 | 4.02 ± 0.84 | 3.75 ± 0.91 | 4.27 ± 0.69 | 4.07 ± 0.87 | 4.30 ± 0.63 |
| | MSHB | 4.35 ± 0.75 (p = 0.397) | 4.31 ± 0.67 (p = 0.095) | 4.33 ± 0.57 (p = 0.020) | 4.29 ± 0.71 (p = 0.908) | 4.29 ± 0.81 (p = 0.290) | 4.81 ± 1.04 (p = 0.119) |
| | LSHB | 3.79 ± 0.46 P = 0.034 | 3.70 ± 0.71 (p = 0.144) | 3.87 ± 0.77 (p = 0.849) | 3.77 ± 0.51 (p = 0.034) | 3.77 ± 0.65 (p = 0.209) | 4.25 ± 0.93 (p = 0.919) |
| | HSLB | 2.47 ± 0.53 (p < 0.001*) | 2.40 ± 0.56 (p < 0.001*) | 2.32 ± 0.49 (p < 0.001*) | 2.41 ± 0.64 (p < 0.001*) | 2.36 ± 0.52 (p < 0.001*) | 2.71 ± 0.71 (p<0.001*) |
| MSHB | | 4.35 ± 0.75 | 4.31 ± 0.67 | 4.33 ± 0.57 | 4.29 ± 0.71 | 4.29 ± 0.81 | 4.81 ± 1.04 |
| | LSHB | 3.79 ± 0.46 0.017 | 3.70 ± 0.71 (p = 0.007) | 3.87 ± 0.77 (p = 0.023) | 3.77 ± 0.51 (p = 0.008) | 3.77 ± 0.65 (p = 0.016) | 4.25 ± 0.93 (p = 0.013) |
| | HSLB | 2.47 ± 0.53 (p < 0.001*) | 2.40 ± 0.56 (p < 0.001*) | 2.32 ± 0.49 (p < 0.001*) | 2.41 ± 0.64 (p < 0.001*) | 2.36 ± 0.52 (p < 0.001*) | 2.71 ± 0.71 (p<0.001*) |
| LSHB | | 3.79 ± 0.46 | 3.70 ± 0.71 | 3.87 ± 0.77 | 3.77 ± 0.51 | 3.77 ± 0.65 | 4.25 ± 0.93 |
| | HSLB | 2.47 ± 0.53 (p < 0.001*) | 2.40 ± 0.56 (p < 0.001*) | 2.32 ± 0.49 (p < 0.001*) | 2.41 ± 0.64 (p < 0.001*) | 2.36 ± 0.52 (p < 0.001*) | 2.71 ± 0.71 (p<0.001*) |

This result was also highlighted by the post-hoc analysis performed among the targets, which revealed that only for right-handers the mean force applied by the *left hand* was significantly larger for the right target than for the left and central ones, in the case of the softer materials (MSHB and LSHB: p<0.001). No differences were showed between C and L target and for LSHB and HSHB materials (p>0.003). The same analysis was conducted for the right hand: target direction had no effects on the force applied during manipulation of the VO (p>0.006), since the dominant hand modulates the force application correctly. No statistical differences among targets have been found for the left-handers (p>0.01), confirming a better control of the force applied between the two hands when manipulating a shared virtual object of different rendered mechanical properties.

The switch of the functional hands' roles depending on the movement direction in right-handers was further highlighted by the significant differences founded between the four analyzed materials. While for the *right hand*, for all target positions, the force' analysis showed a higher mean force in HSHB, MSHB, LSHB materials than HSLB material, for the *left hand* no significant differences were found when subjects were moving towards the right target (see Table II). This result means that the non-dominant hand, which assumes the leading role in the right direction, lacks the ability to discriminate and therefore, to modulate the force when materials of different physics properties are manipulated. Again, in left-handers the statistical analysis between the four haptic features confirmed a better coordination than right handers. In fact, for all the target positions the force' analysis showed a higher mean force in HSHB, MSHB, LSHB materials than HSLB material, for both the *right and left hand* (see Table III). This result means that both dominant and non-dominant hand, are able to discriminate and therefore, to modulate the force when materials of different physics properties are manipulated.

In addition to the analysis carried out when the task was successful, i.e., when the subject reached the target, we evaluated what happened if participants overcame the force and broke the object through the ***Force Percentage Break – FPB***. These further results confirmed the inability of the non-dominant hand to modulate the force. Overall, we found that regardless of direction, the hand that exceeded the force limit was the left (LH). This evidence was valid for all materials except for HSLB and HSHB (rigid materials without compliance) along the left direction (L), Figure 7B. For this metric, it was not possible to perform a statistical analysis since the number of repetitions was not consistent (not all subjects exceeded the required force).

C. *Characterization of movement symmetry and smoothness*

The results related to the ***Dynamical Symmetry Index – DSI*** - indicated that the movement direction influenced the motor control of bimanual action, generating different asymmetries between the two hands. The ***DSI*** trend, depicted in Figure 8A for a single subject and in Figure 8B for the population, was highly influenced by the target position and the VO's typology. When moving towards the left portion of the workspace (L), subjects performed a longer path with the right hand (***DSI*** values between 10 and 20 %) for HSHB, MSHB and LSHB materials. Instead, for the HSLB material, subjects showed an initial negative peak of ***DSI***, for a movement percentage between 0 and 20 % before stabilizing around zero throughout the following movement phase. When the target was presented on the right (R), for all materials, subjects performed a longer path with the left hand (***DSI*** values between -10 and -20 %). Finally, for the central target (C) subjects showed a symmetric behaviour for all the materials and the whole movement phase.

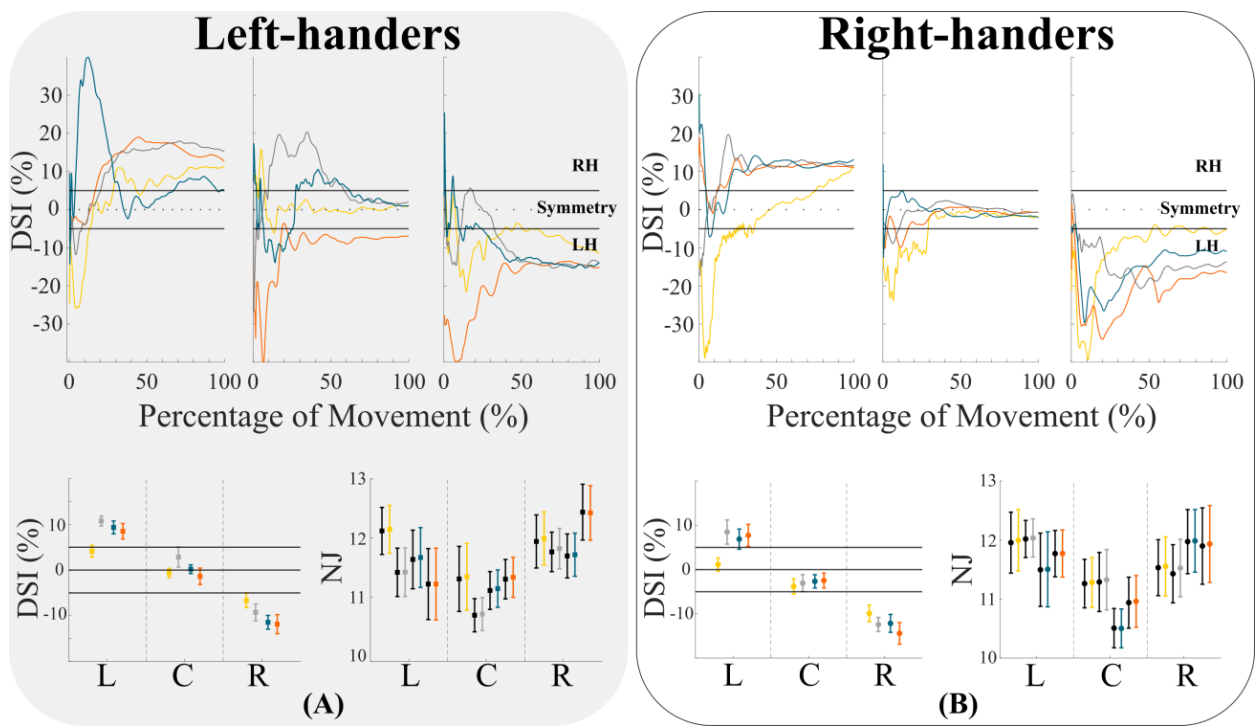

Fig.8. **(A)** Left-handers panel. From top line: Dynamical Symmetry Index – DSI – profiles from a single participant (L, C and R target directions). On the left bottom line: mean and SE values of DSI computed during forward movement across subjects. On the right bottom line: Normalised Jerk -NJ- Mean and SE values computed across participants. Every metrics is represented across the three different targets and for each simulated virtual object. **(B)** Right-handers panel.

The statistical analysis confirmed the previous results, showing a significant difference between the targets ('*Target*' effect: F=84.447, p<0.001), the Handedness (*'Handedness'* effect: F=6.907, p=0.027), an interaction effect between the haptic features and the targets ('*Material * Target*': F=8.083, p<0.001), the Handedness and the targets (*'Handedness\*Target'* effect: F=84.447, p<0.001), and the interaction between Handedness, the haptic feature and the targets (*'Handedness\*Target\*Material'* effect: F=8.083, p<0.001). From the post-hoc analysis, we found in right handers, significant differences between the HSLB and all the other materials for the left target (post-hoc analysis: p<0.001), Figure 8C. No statistical differences between materials were found for the central and right target (p>0.03). In left handers, post hoc analysis showed significant differences between the HSLB and all the other materials when moving both to the right and the left direction (p<0.001).

We evaluated the smoothness of the trajectory employing the ***Normalised Jerk - NJ***. We found lower values and thus a good smoothness, for the central target (C), only for the medium stiffness VO (MSHB) that results in an optimal ratio between amplitude and frequency of deformation oscillations. The statistical analysis evidenced an effect of the direction ('*Target*':

F=16.26, p<0.001) and an effect of the hand ('*Hand*': F= 85.941, p<0.001). We also obtained two interaction effects (*'Material\*Target'*: F=2.44, p=0.032, *'Material\*Hand'*: F=4.57, p=0.027) that allowed us to establish that the direction of movement and the body's side had an effect dependent on the VO proprieties.

In particular, in right handers we found a significant difference between targets only for the MSHB for both the right hand (p=0.001, C (10.51±0.82N) versus R (11.99±1.30N)) and the left hand (p=0.001: C (10.51±0.8N) versus R (11.97±1.33N); p=0.002: C (10.51±0.8N) versus L (11.50±1.52N). For all other materials, no significant differences have been observed (p>0.003). In left handers, post-hoc analysis showed a significant difference between targets only for the HSHB for both the right hand (p=0.002, C (11.15±0.75N) versus R (12.04±0.92N)) and the left hand (p=0.002, C (11.13±0.75N) versus R (12.00±0.75N)). For all other materials, no significant differences have been observed (p>0.015). Furthermore, in right handers we found significant differences between hands only for the HSHB along with the right target (p<0.001) Figure 8D. No statistical differences between materials were found (p>0.213). In left handers no significant differences between hands were found (p>0.07), indicating a better coordination between hands.

*D. The low breakage point highly influences participants' performance in terms of execution time and failures*

As expected, thanks to the quality of the simulation, subjects' performance decreased when manipulating the more fragile material, which was handled with care, and therefore participants needed a longer execution time to succeed. In fact, the statistical analysis on the **Execution Time** parameter showed a significant difference between materials (*'Material'* effect: F=17.111, p<0.001), targets (*'Target'* effect: F=5.542, p=0.013), their interaction (*'Material*Target'* effect: F=2.713, p=0.022), the interaction between handedness and target (*'Handedness*Target'* effect: F=5.542, p=0.013), and the interaction between handedness, target and haptic feature (*'Handedness*Target*Material'* effect: F=2.713, p=0.022).

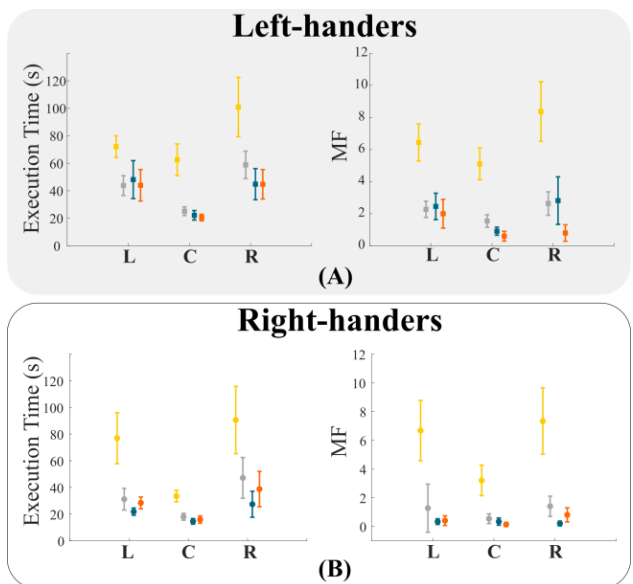

Fig.9. **(A)** Left-handers panel. Execution Time (left side) and Max Failure – MF (right side). Each parameter is represented for each material and for each target with mean and SE values. **(B)** Right-handers panel.

The post-hoc analysis for both right and left handers revealed a significant difference between the HSLB material and all other materials by analysing each direction individually (right handers: L direction: p=0.003 for the comparison HSLB versus HSHB and p=0.001 for the comparison HSLB versus MSHB and HSLB versus LSHB; C direction: p<0.001 for all; R direction: p=0.003 for the comparison HSLB versus HSHB, p<0.001 for HSLB versus MSHB and p=0.003 for HSLB versus LSHB; left handers: L direction: p=0.003 for the comparison HSLB versus HSHB, p=0.001 for the comparison HSLB versus MSHB and p=0.003 for HSLB versus LSHB; C direction: p=0.003 for all; R direction: p=0.003 for the comparison HSLB versus HSHB, p=0.001 for HSLB versus MSHB and p=0.002 for HSLB versus LSHB).

Although the HSLB consisted of high stiffness, the low breakage point led subjects to perform the task slowly and carefully, with subsequent larger **Execution Time**, as showed in Figure 9A.

The Bonferroni corrected t-tests showed also statistical differences in right handers between the targets, highlighting how the direction of movement influenced the motor control in bimanual actions, modulating the **Execution Time** which resulted larger towards the right direction for most of the VOs: HSLB (L versus C: p=0.002), MSHB (L versus C: p<0.001) and LSHB (L versus C: p<0.001). No statistical differences among targets have been found for the left-handers (p>0.01), highlighting that the direction of movements did not influence the bimanual motor strategies in left-handed people.

Finally, the difficulties in performing the task for a given type of simulated VO were analysed by the **Max Failure - MF** - parameter. This latter considered the number of times in which subjects exceeded the maximum force allowed. We had significant differences between handedness (*'Handedness'* effect F=7.007, p=0.027), materials ('*Material*' effect: F= 33.274, p<0.001), and targets ('*Target*' effect: F=3.818, p=0.04). We found for both right and left handers that the difficulties in performing the task increased when the compliance and the force necessary to break the VO decreased: a significant difference between the HSLB material and all other materials has been found by analysing each direction individually (right handers: L direction: p<0.001 for the comparison HSLB versus MSHB and HSLB versus LSHB; C direction: p=0.001 for the comparison HSLB versus HSHB and HSLB versus MSHB, p<0.001 for HSLB versus LSHB; R direction: p=0.001 for the comparison HSLB versus HSHB and p<0.001 for HSLB versus MSHB and HSLB versus LSHB; left handers: L direction: p=0.003 for the comparison HSLB versus MSHB and HSLB versus LSHB; C direction: p=0.001 for the comparison HSLB versus HSHB and HSLB versus MSHB, p<0.001 for HSLB versus LSHB; R direction: p=0.002 for the comparison HSLB versus HSHB and p<0.001 for HSLB versus MSHB and HSLB versus LSHB), Figure 9B. No statistical differences were found between target directions (p>0.03).

## VII. Discussion

A fully immersive virtual scenario must involve all the somatosensory system, including vision, auditory and haptic sense. Therefore, considering the most recently haptic technologies, we decided to investigate the motor strategies during a bimanual reaching task in which participants manipulated a 3D virtual object with four different *haptic features*.

Was the combination of a robotic exoskeleton integrated with VR and haptic interfaces enough to answer our question? How does handedness influence motor coordination in bimanual tasks when virtual objects of different haptic features are manipulated?
Outcomes revealed multiple aspects, which, to our knowledge, have received less attention in previously published contributions, for the reason that most of the literature on bimanual actions primarily focused on *uncoupled* tasks [18], [19], [20], while few contributions specifically focused on bimanual tasks in *coupled* settings [21], [22]. Another reason for this lack of results is the affordability of complex haptic devices, which must be designed to provide robust and accurate force feedback in a three-dimensional workspace and to involve the whole upper limbs.

The coordinated coupled cooperation among upper limbs to achieve a common motor goal is a distinctive feature of human behaviours since individuals use both hands to haptically explore and manipulate the objects in daily actions. In the last decades, bimanual coordination has been the critical planer of intensive investigation concerning how information from the proprioceptive and tactile senses is integrated [17]. However, little is still known about how the handedness influences the different roles of the hands when directional changes occur between body and goal positions. Previous evidence was obtained without the employment of the recent haptics and robotics technologies which provide a broader context to test multiple conditions and simulate various tasks with a high degree of reliability for studying human bimanual manipulation.

Today, the dynamic process involving the use of the two hands can be described by two theories: (i) the coordinated use of limbs is ruled by the dominance hand over the other, which primarily acts, while the other stabilizes the manipulation [26]–[29]; (ii) the hands switch their functional role across various environmental constraints [21], [30]. With this in mind, we wanted to provide further evidence that, by using haptics, the bimanual coordination can be accurately characterized across the human workspace during motor-coupled activities.

### A. Functional hands' roles of right-handers depend on movement direction

The study's central finding was related to the primary acting hand involved in the bimanual object manipulation. For right handers, we identified a significant effect of the movement direction. In particular, we detected a predominance of the non-dominant hand for movements towards the right direction. In fact, the mean force profiles, computed in our study, highlighted a significant asymmetry between hands only for movements towards the right way. Since the virtual objects are not infinitely rigid, the force applied by the left hand is thus propagated within the material without affecting the right hand, or obstructing is movement. For the other directions no differences between the force applied by the two hands were found. This means that in a multi-joint three-dimensional space, the choice of the *acting hand* is related to the type of the investigated task and the movement's desired direction. This result is in line with Contu et al. [13], who showed, at the level of the wrist, that in right-handers subjects, depending on the direction of the movement, for the target towards the right, the leading hand is the non-dominant one, vice versa for the target towards the left direction the leading hand is the dominant one. The explanation can be found in the concept of "*Direction-dependent leading hand*". For clarity, the basis of this phenomenon can be investigated by merely considering right-handed subjects. When movements to the left are required, the leading hand is the dominant one, which in healthy subjects is the one normally used to fine control movements. Vice versa, when dealing with right directions, the guide's role is inverted and transferred on the hand that generally has the function to stabilize, namely the non-dominant one. This determined a significant direction dependent asymmetry: the non-dominant hand, when it plays the role of the leading hand, i.e., in movements directed to the right, it is not able to modulate the force correctly and is inclined to exert more force than the strictly necessary.

The present finding is consistent with the previous studies, which showed that direction is a primary movement parameter coded in various brain structures during unilateral upper limb movements [55], [56].

When right-handed participants moved towards central targets, subjects showed symmetric behavior between the dominant and non-dominant hands.

### B. Left-handers develop a more coordinated control of their non-dominant arm

Left-handers present a more coordinated force control when moving within the workspace: no statistical differences between the forces applied by the two hands have been obtained in all the target directions. This result agrees with Przybyla et al. [57], who showed how the amplitude of the interlimb differences in terms of movement curvature, accuracy, and precision were substantially smaller in left- than right-handers during unimanual reaching tasks. Such findings are consistent with previous studies [58]-[59], which have also reported a reduced lateralization in left-handers during the execution of sequential movements. The reduction in the dominant hand predominance of left-handers could be related to environmental stresses, requiring more elaborate use of the right non-dominant arm over the course of one's life; for example, the scissors shape requires right-hand manipulations, whether one is left- or right-handed.

### C. Compliance perception affects motor performance

Motor performance is influenced by the materials' compliance haptically rendered, as revealed by the outcomes related to the **Execution Time and Max Failure-MF**. Information from different senses is separately processed and converges into a unique environment so that the perception is the best possible estimation [60]. According to Hooke's law, compliance is the combination of position and force information. Since position information is provided from the *haptic* and *visual* modalities, subjects showed the highest performances for the more compliant materials. Indeed, as expected, the time necessary to successfully achieve the task was significantly higher for the material with the lower breakage point, meaning that it was handled with more care. Also, the MF parameter showed that the difficulties in the task execution increased when the compliance and the force necessary to break the virtual object were decreased, meaning that the most fragile material decreased user performance.

### D. Limitations

Our study presents anyway limitations: the first concerns the absence of an HMD (head-mounted display**)** device for providing users with an immersive environment. Such technology could lead to different outcomes compared to those obtained in the present experimental scenario since it would elicit a more realistic perception and reaction in the participants. Our interaction environment was rendered on a flat 2D screen, resulting in a possible altered estimation of the movement-planning against the targets. In future studies, the inclusion of an immersive virtual reality environment may enhance the evidence found in the current study.

Another limitation relates to the inclusion of only proximal actuated joints: the *shoulder joints* (sh-AA, sh-PS and sh-FE) and the *elbow joints* (eb-FE). During the execution of dexterous actions, also the distal joints, like the wrist, are recruited. In the framework of the current study, the wrist flexion/extension joint was blocked so that the resultant force, provided at the exoskeleton's handle, was distributed only proximally. The integration of more degrees of freedom, i.e., by including an actuated wrist exoskeleton [61], would involve the whole arm kinematic chain, allowing users to perceive more realistic forces. Moreover, in the same perspective of providing the user with haptic sensations as realistic as possible, further future developments would be to realize virtual objects with different shapes, implement more realistic virtual hands in the god-object method, and use an actuated glove [62]. In such a way, the participant could interact with the object's surface directly without being constrained to the use of a handle.

VIII. CONCLUSION

This study aims to provide a broader and more comprehensive view of the motor control strategies in bimanual coupled manipulation across a three-dimensional workspace through the use of the ALEx-RS robotic exoskeleton, integrated for the first time with haptic feedback.
The adoption of such a device, which provides multimodal sensory feedback, showed that haptic and visual feedback might influence the role of dominant and non-dominant hands during the dynamic coupled bimanual task. Results also indicated that manipulating an object with higher compliance improves task performance.
Our outcomes on healthy subjects (both right-handed and left-handed) were aimed at showing the potentialities of the implemented haptic interface since the designed control represents a starting point for a fully customized and measurable haptic environment. The current setup implementation could have implications and usages for several applications such as clinical assessments and rehabilitation treatments in a highly controlled but, most importantly, safe, and measurable environment that strives to mimic real-world conditions as reliably as possible.

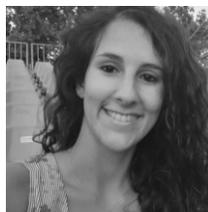

**Elisa Galofaro** is a Postdoc Researcher at Heidelberg University (Germany) at the Institute of Computer Engineering (ZITI), under the supervision of Professor Lorenzo Masia. She received her bachelor's degree in biomedical engineering (September 2014) and her master's degree in Bioengineering (curriculum Neuroengineering and bio-ICT) both from the University of Genoa (February 2017). In November 2017 she started her PhD in Bioengineering and Bioelectronics at the University of Genoa under the supervision of Professor Maura Casadio. Her main research interests are in rehabilitation robotics, haptics interaction, proprioception and assistive technologies.

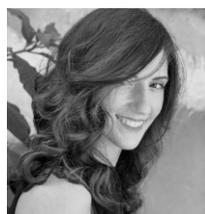

**Erika D'Antonio** is a Postdoc Researcher at Heidelberg University (Germany) at the Institute of Computer Engineering (ZITI), under the supervision of Professor Lorenzo Masia. She received the bachelor's degree cum laude in biomedical engineering from University of Roma "La Sapienza", in January 2014. She took the master's degree cum laude in Biomedical Engineering in January 2016. In November 2017 she started her PhD in Industrial Engineering at the Niccolò Cusano University of Rome, under the supervision of Professor Fabrizio Patané. Her main research activity focuses on the development of

 

motion-analysis systems, robotic mechanisms, exoskeletons and virtual/augmented reality setups for neurorehabilitation.

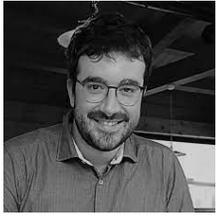

**Nicola Lotti** is a Postdoc Researcher at Heidelberg University (Germany) at the Institute of Computer Engineering (ZITI), under the supervision of Professor Lorenzo Masia. He received his bachelor's degree in biomedical engineering (July 2014) from University of Bologna and his master's degree in Bioengineering (curriculum Neuroengineering and bio-ICT) from the University of Genoa (October 2016). In November 2016 he started his PhD in Bioengineering and Bioelectronics at the University of Genoa working at the Department of Informatics, Bioengineering, Robotics, and System Engineering under the supervision of Professor Vittorio Sanguneti. His main research interests are in motor control, rehabilitation robotics, exoskeleton, computational model.

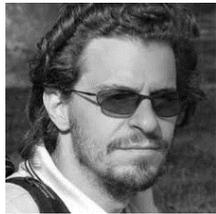

**Fabrizio Patané** is an associate professor of Thermal and Mechanical Measurements with the Faculty of Engineering at the "Niccolò Cusano" University, Rome, Italy. He received Mechanical Engineering degree at "La Sapienza" University of Rome in 2000, and PhD Degree at Padova University in 2004. At the "Niccolò Cusano" University, he is coordinator of both the Teaching and the Research Commissions and is member of the doctorate school teaching staff in Industrial-Civil Engineering. His scientific skills regard mechanical and thermal measurements, environmental and inertial measurements, motion-analysis systems for biomedical applications and robotic mechanisms for neurorehabilitation.

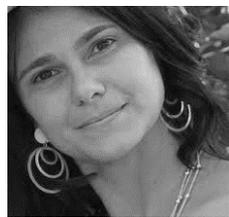

**Maura Casadio** is an Associate Professor of Biomedical Engineering at the University of Genoa, Italy. She received a Master's degree in Electronic Engineering (2002) at the University of Pisa, a Master's degree in Bioengineering (2007) and a PhD in Robotics and Bioengineering (2006), both at the University of Genoa. She has been working (2008-2011) as postdoctoral fellow in the Department of Physiology, Northwestern University and the Robotics Laboratory, Rehabilitation Institute of Chicago. Her main areas of interest are neural control of movement, robots for rehabilitation and body-machine interfaces.

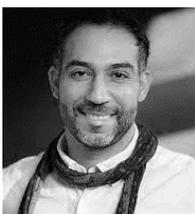

**Lorenzo Masia** is Full Professor in Medical Technology and Biorobotics at Heidelberg University (Germany) at the Institute of Computer Engineering or Institut für Technische Informatik (ZITI). He graduated in Mechanical Engineering at "Sapienza" University of Rome in 2003 and in 2007 he accomplished his PhD at the University of Padua. He was at first postdoctoral researcher at the Italian Institute of Technology (IIT) and then Team Leader of the Motor Learning and Rehabilitation Laboratory of the Robotics Brain and Cognitive Sciences Department. He was an Assistant Professor at the School of Mechanical & Aerospace Engineering (MAE) at Nanyang Technological University (NTU) of Singapore, leading the ARIES Lab (Assistive Robotics and Interactive Ergonomic Systems); He was Associate Professor in Biodesign at University of Twente (The Netherlands) from June 2018 to March 2019.